\documentclass[10pt,twocolumn,letterpaper]{article}

\usepackage{wacv}              

\usepackage{graphicx}
\usepackage{amsmath}
\usepackage{amssymb}
\usepackage{booktabs}
\usepackage{float}

\usepackage{xcolor}
\usepackage{multirow}
\usepackage{colortbl}
\usepackage{adjustbox}
\usepackage[pagebackref,breaklinks,colorlinks]{hyperref}
\usepackage[capitalize]{cleveref}
\crefname{section}{Sec.}{Secs.}
\Crefname{section}{Section}{Sections}
\Crefname{table}{Table}{Tables}
\crefname{table}{Tab.}{Tabs.}

\def\wacvPaperID{1219} 
\def\confName{WACV}
\def\confYear{2025}

\title{Can Out-of-Domain data help to Learn Domain-Specific Prompts for Multimodal Misinformation Detection?}

\author{Amartya Bhattacharya$^{1*\dagger}$,
    Debarshi Brahma$^{2*}$,
    Suraj Nagaje$^{2*}$, \\
    Anmol Asati$^{2}$, 
    Vikas Verma$^{2}$,
    Soma Biswas$^{2}$ }

\begin{document}





\twocolumn[
\maketitle
\vspace{-1.5em} 
\begin{center}
    $^{1}$BBI Inc., $^{2}$Indian Institute of Science, Bangalore \\
    {\tt\small amartya.bhattacharya@bbiinc.ai, \\ \{debarshib, surajnagaje, anmolasati, vikasverma, somabiswas\}@iisc.ac.in}
\end{center}
\vspace{2em} 
]

\def\thefootnote{*}\footnotetext{Equal contribution.}
\def\thefootnote{$\dagger$}\footnotetext{Work done at the Indian Institute of Science, Bangalore.}

\begin{abstract}
Spread of fake news using out-of-context images and captions has become widespread in this era of information overload. Since fake news can belong to different domains like politics, sports, etc. with their unique characteristics, inference on a test image-caption pair is contingent on how well the model has been trained on similar data. 
Since training individual models for each domain is not practical, we propose a novel framework termed DPOD (Domain-specific Prompt tuning using Out-of-domain data), which can exploit out-of-domain data during training to improve fake news detection of all desired domains simultaneously.
First, to compute generalizable features, we modify the Vision-Language Model, CLIP to extract features that helps to align the representations of the images and corresponding captions of both the in-domain and out-of-domain data in a label-aware manner. 
Further, we propose a domain-specific prompt learning technique which leverages training samples of all the available domains based on the extent they can be useful to the desired domain. 
Extensive experiments on the large-scale NewsCLIPpings and VERITE  benchmarks demonstrate that DPOD achieves state of-the-art performance for this challenging task. Code: \color{magenta}https://github.com/scviab/DPOD\color{black}.
\end{abstract}


\section{Introduction}
\label{sec:intro}

In this digital world, social media has become the main source of information for an increasing fraction of the population. Thus, it is very important that the news reaching the masses is authentic, since fake news can have serious consequences, like manipulating public opinion, stirring up conflicts, and even impacting financial markets. 
One increasingly common form of fake news is the out-of-context use of images, where a real image is paired with a false caption before dissemination. It is very difficult to detect such fake news, since the image is real, i.e., not manipulated or generated, and it is only recently that researchers have started to address this challenging problem \cite{cosmos}\cite{open_domain}.

In real-world, fake news can belong to any domain like {\em healthcare, politics, entertainment}, etc. Obviously, the accuracy of authentication of a test image-text pair belonging to a particular domain depends on how well that domain is represented in the training data. 
Training different models for different domains is not a practical solution, since it assumes abundant availability of data from every domain, in addition to not effectively utilizing the inter-relations and overlapping traits that may be present among them. 

Towards this goal, we propose a novel, generalized framework termed DPOD ({\bf D}omain-specific {\bf P}rompt-tuning using {\bf O}ut-of-{\bf D}omain data) for out-of-context misinformation detection, also termed as Multi-modal Fake News Detection (MFND). 
The model is trained using data from all the available domains, such that it can then be simultaneously tailored to each individual domain, by leveraging information from all the other domains.
As an example, for MFND on a particular domain of interest like {\em politics}, we effectively utilize out-of-domain data from {\em healthcare}, {\em international relations}, etc. to improve its performance.
The DPOD framework first utilizes label-aware alignment of the available image-text pairs of all the domains to learn generalizable features, irrespective of the domain of interest.
Here, we harness the capabilities of a large Vision Language Model (CLIP)~\cite{CLIP} to comprehend interwoven text and image features. 
Next, a semantic domain vector is computed for each of the training data domains, which incorporates the similarity of a particular domain with all the other available training domains.
Further, this is utilized to learn a domain-specific prompt, along with a set of generic learnable prompts, which is finally used for predicting whether a given image-caption pair is real or fake. 
During inference, the model specialized to the test domain can  be utilized, even when the test domain is unknown.
Extensive experiments on two benchmark datasets, namely NewsCLIPpings~\cite{NewsCLIPpings} and VERITE~\cite{verite}, demonstrate the effectiveness of DPOD over state-of-the-art approaches. 
The contributions of this work can be summarized as follows: \\
1) We propose a novel framework, termed DPOD, which effectively utilizes out-of-domain data for improving the performance of in-domain data. \\ 
2) We utilize a label-aware loss to learn a generalizable model, and propose a novel framework to learn domain-specific prompts using out-of-domain data. \\
3) This helps in quick model deployment for the target domain, even with limited domain-specific annotated data, thus being effective for scarcely annotated data scenarios. \\
4) Extensive experiments show that the proposed DPOD outperforms the existing approaches in different settings, thus achieving the new state-of-the-art.

Now, we briefly describe the related work, followed by the proposed framework and the experimental evaluation.


\section{Related Work}
Here, we briefly describe the related work in MFND, self-supervised learning and prompt tuning. \\
{\bf Multi-modal Fake News Detection:}
To address the MFND task,~\cite{NewsCLIPpings} proposed NewsCLIPpings dataset, derived from VisualNews~\cite{Visual_news}, and suggested using CLIP and VisualBERT feature extractor.
SAFE~\cite{SAFE} utilizes Text-CNN and VGG \cite{vgg} model for extracting features from the texts and images which are then used to predict the fake news. Similarly, \cite{spotfake} use BERT \cite{bert} and VGG-19 to extract textual and visual features and then fuse them for classification. Most of these  methods fail to perform well on unseen events. \cite{EANN}  trained an event discriminator network concurrently with multi-modal fake news detector to eliminate event-specific characteristics while retaining common features among different events. 
~\cite{cross_domain_FND} jointly preserves domain-specific and cross-domain knowledge in news records to detect fake news from different domains.
In~\cite{Meta_neural_process}, meta-learning and neural process methods are integrated to achieve high performance even on events with limited labeled data.

\begin{figure*}[t!]
  \centering
  
  \includegraphics[height=7.7cm]{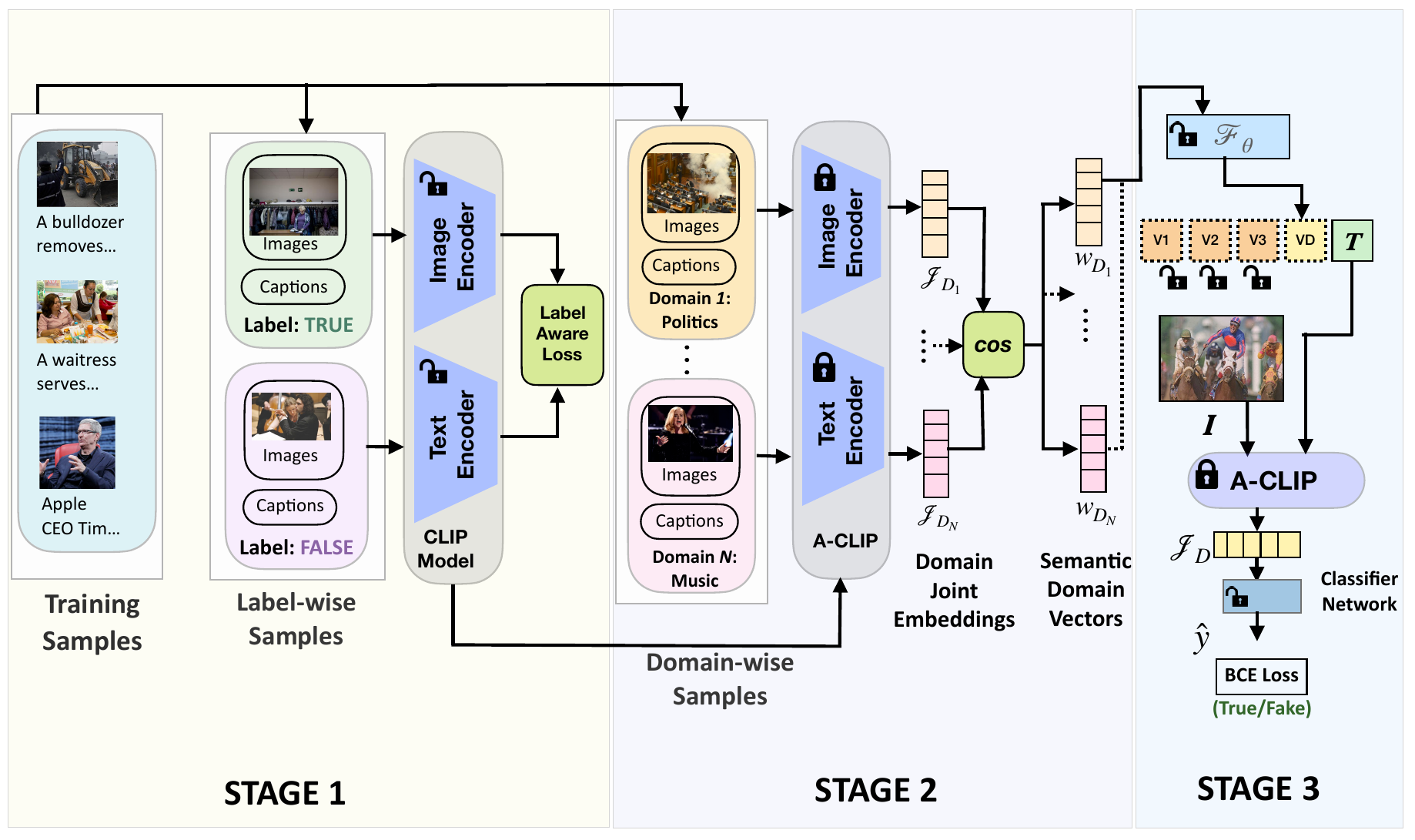}
  \caption{Illustration of the different stages of the proposed  DPOD approach. 
  In Stage 1, we learn the image and text encoders via Label-Aware Alignment Loss. 
  Then this model is used to obtain semantic domain vectors in Stage 2.
  This, in turn is finally used to learn domain-specific and generic prompts with out-of-domain samples to predict the veracity of the news in Stage 3. \color{black}}
  \label{fc}
\end{figure*}
\label{sec: Label aware alignment}

Recently, large scale vision-language models (VLMs) like CLIP~\cite{CLIP} and ALIGN~\cite{Allign}, which learns information from both images and texts, have greatly advanced computer vision tasks. 
\cite{CLIP_guided} used CLIP along with ResNet and BERT to extract  multimodal features and used attention over different modalities to get the final features for classification. 
\cite{papadopoulosacmmad} proposed a novel framework for MFND using CLIP encoders and a Transformer-based detector.
Some recent works also use social media related information to detect  fake news~\cite{cross_domain_FND}\cite{open_domain}\cite{zhangarxiv}. 
\cite{cross_domain_FND} uses image, text and propagation graph for MFND and \cite{zhangarxiv} suggested a graph model based on CLIP embeddings.  
~\cite{open_domain} proposes a Consistency-Checking Network to measure the consistency of the image-text pairs along with evidences collected from the internet.  \cite{ecenet} uses a combination of fine-grained and course-grained attention network along with deep reinforcement learning, while very recently \cite{sniffer} has explored the use of MLLMs in this problem. 
{\em In our work, we do not assume the availability of open-domain information. } \\ \\
\textbf{Self-Supervised Learning (SSL):}{
SSL leverages the inherent structure of data to generate its own supervisory signals without relying on large labeled datasets. \cite{simCLR_paper} have showed that the combination of data augmentations has a crucial impact on predictive tasks. 
Supervised contrastive loss~\cite{supcon} leverages label information for self-supervised representation learning. It pulls together clusters of points belonging to the same class while pushing apart clusters of samples from different classes.
Recently, self-supervision has been used successfully for out-of-context misinformation detection using images and text. 
In~\cite{cosmos}, SSL is used to align individual objects in an image with the corresponding textual claims, where, given two captions, the goal is to detect whether an image is being used out-of-context. 
The state-of-the-art method ~\cite{self_supervised} propose a Self-Supervised Distilled Learner, which uses self-supervised learning to detect fake news, where they propose a Teacher network to guide a Student network in mimicking similar decision patterns.
{\em Though we also utilize self-supervision, our proposed DPOD framework is much simpler, yet more effective, and does not rely on teacher-student model.}
}\\ \\
\textbf{Vision-Language Models and Prompt Tuning:} 
Adapting large VLMs like CLIP to downstream tasks is challenging. Parameter efficient finetuning techniques like prompt learning is a popular approach, where the text prompt tokens are learned during the finetuning stage.
Recently, many works have proposed prompt learning techniques such as CoOp \cite{coop}, CoCoOp~\cite{cocoop}, Maple~\cite{maple}, etc., which increases the generalizability of the VLMs. Some works specifically address the domain adaptation problem, e.g., \cite{dom_inv_prompt} proposes domain-invariant prompt tuning, which can generalize to unseen domains. \cite{switchprompt} uses unimodal prompt tuning (in language models), whereas, \cite{carte} uses only visual prompt tuning.  The works in~\cite{dapl,adclip,poda} learns prompts specifically for each domain. 
{\em In this work, we explore how prompt learning can be used for the challenging MFND problem.}

\section{Problem Definition and Model Overview}
\noindent
The goal is to develop a generalized MFND model, which effectively incorporates the  domain information of the training data and relation between the different domains during training.
Let the train dataset be denoted as $\mathcal{S}_{train}=\{{D}_{1},{D}_{2},...,{D}_{n}\}$. ${D}_{i}$ denotes each domain and can be written as, ${D}_{i} = \{({I}_{k}, {T}_{k}, {y}_{k})\}, k = 1,2,...,{n}_{{D}_{i}}$, where ${n}_{{D}_{i}}$ is the number of samples in domain ${D}_{i}$. ${I}_{k}$ and ${T}_{k}$ denote the $k^{th}$ image and text pair, and ${y}_{k}$ denotes the ground truth, i.e., fake or true news. 
The news samples can originate from various sources
(or domains) such as politics, healthcare, sports, etc. 
We assume that we know the domain of the training examples, though we show empirically that the proposed framework can inherently handle domain label inconsistencies that is expected from crowd-sourced data annotation. 
The model is trained once using all the input domains, such that depending on the domain of the test image-text pair $D \in \{{D}_{1},{D}_{2},...,{D}_{n}\}$, the domain specific model is utilized for best performance. 
First, we briefly describe the base model used. \\ 

\noindent{{\bf Base Architecture.}}
Recently, vision language foundation models like CLIP \cite{CLIP}, Align \cite{Allign} etc., have been immensely successful in relating images and the corresponding text and have also been used for the MFND task~\cite{self_supervised},~\cite{open_domain}. 
CLIP is trained using natural language, which being expressive, can supervise a much broader set of visual concepts. 
Following the recent literature, we also use a modified CLIP architecture for our work.
The broader structure of our proposed DPOD framework can be divided into three stages as follows (Fig.~\ref{fc}):\\
{\bf Stage 1: Label-Aware Alignment of Multi-modal data:} In this stage, we train the encoders of the original CLIP model in an end-to-end manner using the label-aware alignment loss on  the training data from all domains, to obtain the aligned clip, termed here as A-CLIP. \\
{\bf Stage 2: Creating Semantic Domain Vectors.} Here, we freeze the already learnt image and text encoders of the A-CLIP model. 
Using the domain-wise joint image-text embeddings from these frozen encoders, we create semantic domain vectors for each of the training domains. 
Each index of this vector denotes its similarity with the joint embedding of all the other domains. \\
{\bf Stage 3: Domain-Specific Prompt tuning:} With the above settings, we next propose a prompting strategy during training. Following previous works \cite{coop}\cite{cocoop}, we append a prefix prompt to the text caption. However, a fixed prompt like ``A photo of'' \cite{CLIP}, may not be the best prefix for all the captions. 
Hence, we make the prefix prompt tokens learnable as $\{{V}_{1}, {V}_{2}, {V}_{3}\}$. 
Additionally, we pass the semantic domain vector though a linear layer $\mathcal{F}_{\theta}$, and get a domain-specific prompt token, which is appended to the prefix prompt tokens. The complete text captions with the learnable prompts, and the corresponding images are passed though the frozen encoders to create the final joint embedding, which is subsequently classified into fake or real news.

\begin{figure}[t]
    \centering
        \includegraphics[height=3.7cm]{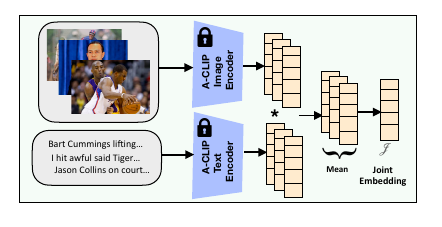}
        \caption{ Computation of the joint embedding of the images and text for a particular domain. These embedding are used in Stage 2 and 3 of our DPOD framework. }
        \label{fig:joint_embed}

\end{figure}

\section{Proposed DPOD Framework}
Here, we describe each of the three stages of the DPOD framework presented in Fig.~\ref{fc} in detail.
\subsection{Stage 1: Label-Aware Alignment}
  

If we want to predict the veracity of news (image-caption pair) in a particular domain ($D$, say {\em politics}), the ideal scenario is when vast amounts of annotated data for domain $D$ are available. 
In real scenarios, getting large amounts of annotated data from each domain is challenging.
But, there might be overlapping concepts or objects in the data from related domains, like {\em politics, international news}, which can be utilized to bolster the learning for each domain.

Using self-supervision as a warm-up stage for model training has been used successfully in previous works like \cite{cosmos,self_supervised}.
In contrast to these approaches,
here, the original CLIP model is trained to learn relations between texts and images from all the domains, but in a label-aware manner. 
Specifically, we use a multi-modal contrastive loss, such that the backbone model learns to bring closer the augmentations of the image and the corresponding text when the news is real and can separate them when the news is fake. 

For label-aware alignment, initially, four augmentations, namely, color jitter, random crop, horizontal flipping and normalization are generated per image.
Then the CLIP embeddings of the original image, text and the image augmentations are generated. 
Suppose an image and the corresponding text be denoted as $I_k$ and $T_k$ respectively and the image augmentations be denoted by $I_{ak}, a=1,\hdots,4$. 
We denote the corresponding embeddings as $\{\hat{I}_k, \hat{T}_{k}, \hat{I}_{ak}\}$.
For the consistent (real) image-text pairs, we want to bring the embeddings of the image ($\hat{I}_{k}$), its augmentations ($\hat{I}_{ak}, a =1, \hdots,4$) and the corresponding text ($\hat{T}_{k}$) close to one another.
Thus for the positive pair, we have two combinations, namely image-image pair and image-text pair.
The corresponding label-aware alignment losses for the $i^{th}$ image augmentation of $k^{th}$ input sample are calculated as:
\begin{eqnarray}
\nonumber
\mathcal{L^{\textrm{1}, \textrm{true}}_{LA}}(I_{ik},I_{jk}) = -\log ( \frac{\exp(\text{sim}(\hat{I}_{ik}, \hat{I}_{jk}) / \tau)}{\mathcal{D}_{true}}) 
\\
\mathcal{L^{\textrm{2}, \textrm{true}}_{LA}}(I_{ik},T_k) = -\log ( \frac{\exp(\text{sim}(\hat{I}_{ik}, \hat{T}_{k}) / \tau)}{\mathcal{D}_{true}}) 
\label{loss-la}
\end{eqnarray}
where $\hat{I}_{ik}$ and $\hat{I}_{jk}, j \neq i$ denotes the $i^{th}$ and $j^{th}$ augmentations of the image $I_k$. 
$\tau$ is the temperature parameter that controls the smoothness of the probability distribution, $N$ is the number of training image-text pairs and $sim$ represents the cosine similarity.
We use $\tau = 0.05$ for all our experiments.
The denominator $\mathcal{D}_{true}$ represents the combination of all the negative pairs and is given as 



     \vspace{-10pt}

\[
\begin{aligned}
 \mathcal{D}_{true} = \sum_{\substack{t=1\\ t \neq k}}^{N} 
 \biggl(\sum_{a=1}^{4}  \exp(\text{sim}(\hat{I}_{ik},\hat{I}_{at}))+
\exp(\text{sim}(\hat{I}_{ik},\hat{T}_{t}))+\\
\exp(\text{sim}(\hat{T}_{k},\hat{T}_{t}))\biggl)
 \end{aligned}
\]

    \vspace{-6pt}    
The total loss corresponding to the real or true news is obtained by adding the losses in eq.~(\ref{loss-la}):
\begin{equation}
\mathcal{L^{\textrm{true}}_{LA}} = \mathcal{L^{\textrm{1}, \textrm{true}}_{LA}} + \mathcal{L^{\textrm{2}, \textrm{true}}_{LA}}
\end{equation}
For the inconsistent (fake) image-text pair, the label-aware alignment of the images with their augmentations is computed in the same manner as in the true case.
Since the image-text pair is inconsistent, they are pushed apart unlike in the true case. 
Thus the final loss for the fake pair is:
\begin{equation}
\mathcal{L^{\textrm{fake}}_{LA}}(I_{ik},I_{jk}) = -\log ( \frac{\exp(\text{sim}(\hat{I}_{ik}, \hat{I}_{jk}) / \tau)}{\mathcal{D}_{fake}})
\end{equation}
Here, the denominator is given as 


    

\[
\begin{aligned}
 \mathcal{D}_{fake} = \sum_{\substack{t=1\\ t \neq k}}^{N} \biggl(\sum_{a=1}^{4}  \exp(\text{sim}(\hat{I}_{ik},\hat{I}_{at})) +  \exp(\text{sim}(\hat{T}_{k},\hat{T}_{t})+\\
 \exp(\text{sim}(\hat{I}_{ik},\hat{T}_{t}))\biggl) 
 +\exp(\text{sim}(\hat{I}_{ik},\hat{T}_{k}))
 \end{aligned}
\]

The final label-aware alignment loss is computed by taking a weighted sum of the two losses corresponding to real and fake data using the eq.~(\ref{loss}) below
\begin{equation}
\mathcal{L_{LA}} = \beta * \mathcal{L^{\textrm{fake}}_{LA}} +\mathcal{L^{\textrm{true}}_{LA}}
\label{loss}
\end{equation}
The parameter $\beta$ is used to weigh the two terms. 
This differential treatment of fake and real samples in both stages helps to better train the model and reduce confusion. 
\subsection{Stage 2: Computing Semantic Domain Vectors}
Here, we create semantic domain vectors which represent how similar each domain is to all the domains in the training dataset. 
If the number of domains in the training dataset $\mathcal{S}_{train}$ is $n$, we create $n$ semantic domain vectors each having $n$ indices. 
Let us take the \textit{``politics''} domain in the training dataset as an example. 
To compute the semantic domain vector for \textit{``politics''}, we pass all its images and corresponding text captions, i.e., $\{({I}_{k}, {T}_{k})\}, k = 1,2,...,{n}_{{D}_{i}}$ through the frozen encoders of the A-CLIP trained in Stage 1 to get $\{({\hat{I}}_{k}, {\hat{T}}_{k})\}$. Here, ${n}_{{D}_{i}}$ denotes the number of training samples in the politics domain. 
We take a Hadamard product of the image and text embedding $\{({\hat{I}}_{k}, {\hat{T}}_{k})\}$ to get ${n}_{{D}_{i}}$ joint embedding. 
Then, we compute the mean across all the sample joint embeddings to get the final mean joint embedding $\mathcal{J}_{{D}_{i}}$, particular to a domain ${D}_{i}$ (here, \textit{``politics''}) as shown below:


\begin{equation}
\mathcal{J}_{{D}_{i}} = \frac{\sum_{k=1}^{{n}_{{D}_{i}}} \mathcal{J}_{k}}{{n}_{{D}_{i}}},  where,  
\mathcal{J}_{k} = \hat{I}_{k} \odot \hat{T}_{k}
\label{5}
\end{equation}
This is illustrated in Fig.~\ref{fig:joint_embed}.
Finally, to create the semantic domain vector ${w}_{{D}_{i}}$, we take the cosine similarity between this joint embedding $\mathcal{J}_{{D}_{i}}$ and those of the other training domains $\mathcal{J}_{{D}_{k}}, k = 1,2,...,n$. Hence, the $j^{th}$ index of the ${w}_{{D}_{i}}$ vector will be the cosine similarity of $\mathcal{J}_{{D}_{i}}$ with $\mathcal{J}_{{D}_{j}}$, ($j \in k$) as shown in eq.~(\ref{7}).
\begin{equation}
{{w}_{{D}_{i}}(j)} = \frac{{\mathcal{J}_{{D}_{i}} \cdot \mathcal{J}_{{D}_{j}}}}{\| \mathcal{J}_{{D}_{i}} \| \| \mathcal{J}_{{D}_{j}}\|}
\label{7}
\end{equation}
We term these as semantic domain vectors, since they can implicitly capture the semantics of the domains, i.e. the domain vectors of similar domains are closer compared to those of unrelated domains. This is illustrated in detail later.

\subsection{Stage 3: Domain-Specific Prompt-tuning with Out-of-Domain data}

Next, we utilize the semantic domain vectors to learn domain-specific prompt embedding that help to incorporate domain-specific information in the model while classifying a news item as true or fake. We initialize a prompt prefix with the standard text \textit{``A photo of''}~\cite{CLIP}. However, we make this prefix learnable (similar to \cite{coop}), since it may not be the best choice for our application. 
We denote these generic learnable prompts as $\{{V}_{1}, {V}_{2}, {V}_{3}\}$. 
To incorporate the domain information, we pass the corresponding semantic domain vector ${w}_{{D}_{i}}$ (obtained in the previous stage) through a linear layer, denoted by $\mathcal{F}_{\theta}$. This layer $\mathcal{F}_{\theta}$ projects the ${n}_{{D}_{i}}$-dimensional domain vector to a 512-dimensional domain prompt token $\mathcal{V}_{{D}_{i}}$, which is then appended to the prompts $\{{V}_{1}, {V}_{2}, {V}_{3}\}$. This forms the whole learnable prompt structure, where $\{{V}_{1}, {V}_{2}, {V}_{3}\}$ is trained in a domain-agnostic manner, whereas, $\mathcal{V}_{{D}_{i}}$ is trained by  conditioning on the domain of the particular training sample. 
This entire prompt is then prepended to the text caption ${T}$ and passed through the frozen A-CLIP text encoder to get a text embedding $\mathcal{\hat{T}}$ as shown in eq. (\ref{8}).

\begin{equation}
    \mathcal{\hat{T}} = \{\theta_{1}, \theta_{2}, \theta_{3}, \mathcal{F}_{\theta}({w}_{D}),  T\}
\label{8}
\end{equation}
Here, $\theta_{1}, \theta_{2}, \theta_{3}$ denotes the learnable parameters for the prefix prompts.
Similarly, the image is passed through the frozen A-CLIP image encoder to get the image embedding $\mathcal{\hat{I}}$. We again create a joint embedding $\mathcal{\hat{J}}$ by taking a hadamard product of $\mathcal{\hat{T}}$ and $\mathcal{\hat{I}}$, and pass it through a Classifier Network.

The Classifier Network consists of two fully connected layers which projects $\mathcal{\hat{J}}$ to $\mathcal{\hat{X}}$. Inspired by \cite{CLIP-adapter}, we also introduce a residual connection from $\mathcal{\hat{J}}$ to $\mathcal{\hat{X}}$, the output of which finally passes through two fully connected layers to give a final prediction $\hat{y}$. The whole network is trained in an end-to-end manner using a Binary Cross-Entropy loss as shown in eq.~(\ref{9}).

\begin{equation}
    \mathcal{L(\theta}) = -{(y\log(\hat{y}) + (1 - y)\log(1 - \hat{y}))}
\label{9}
\end{equation}

\noindent\ Here, $\theta$ denotes all the learnable parameters in the network. 
The ground-truth ${y} = 1$, if the sample is fake, and ${y} = 0$, if it is a true news. \\ 
\noindent\textbf{Inference.} During inference, we consider two scenarios: one where we have the domain information for the test image-text pair, and another where the target domain is unknown. For the first case, we feed its corresponding semantic domain vector through the already trained $\mathcal{F}_{\theta}$ network to get the target specific domain prompt ${V}_{D}$. For the second case, we find the most similar domain of the image-text sample by comparing its joint embedding with all the domain-specific mean joint embeddings, since the domain is unknown. Specifically, we find the domain $D_t$ of this test pair as follows: $D_t = \operatorname*{argmax}_{D} (\mathcal{J}_t \cdot \mathcal{J}_{D_i}), i=1,2,...,n$, where, $\mathcal{J}_t$ denotes the joint embedding of the test sample, and $\mathcal{J}_{D_i}$ denotes the domain-wise mean joint embeddings.
Next, the text caption (of test data) is appended with the learnt prompt vectors to get $\{{V}_{1}, {V}_{2},{V}_{3}, {V}_{D}\}$. 
The final text and image embedding obtained using the frozen A-CLIP encoders are then used to compute the joint embedding. The classifier network gives the final prediction ${\hat{y}}$, which is thresholded to give a binary prediction as True or Fake. \\ \\
\noindent\textbf{Difference with other prompt-learning techniques.} As discussed in the Related Work section, there are some existing works on domain-based prompt tuning~\cite{dapl,adclip}.
But there are few important differences between these techniques and the proposed one: \\
{\bf 1)} In the existing works,  \textit{``domain''} or \textit{``style''} refers to changes in image features. 
Contrary to this, in our case, domains (such as, ``politics'', ``law-crime'', etc.) do not have a significant change in image features (since, images from multiple domains will contain similar objects like humans, etc.), but are subjective to human interpretation of both the images and the texts jointly. \\
{\bf 2)} In existing works, the text  only contains the class name as opposed to a sentence describing the image as in ours. \\
{\bf 3)} We propose a novel strategy of creating semantic domain vectors, which leverages similarity from out-of-domain data, and utilize it to create domain-specific learnable prompts. To the best of our knowledge, this type of domain-prompting strategy has not been explored earlier.

\setlength{\tabcolsep}{8pt}
\begin{table}[tb]
    \caption{Performance comparison of the proposed DPOD framework with existing approaches on the NewsCLIPpings dataset, where we train and test on all the domains.}
    \centering
    \begin{adjustbox}{max width = \linewidth}
    \scalebox{0.85}{
    \begin{tabular}{lcc}
        \toprule
        \textbf{Methods} & \textbf{Backbone} & \textbf{Accuracy (\%)} \\
        \midrule
        DT-Transformer \cite{papadopoulosacmmad} & CLIP ViT-B/32 & 65.67 \\
        MNSL \cite{zhangarxiv} & CLIP & 68.20 \\
        DCD \cite{huangicassp} & CLIP & 65.40 \\
        \midrule
        \multirow{2}{*}{NewsCLIPpings \cite{NewsCLIPpings}} & CLIP ViT-B/32 & 60.23 \\
        & CLIP-RN50 & 61.62 \\
        \midrule
        \multirow{2}{*}{CLIP-FT \cite{open_domain}} & CLIP ViT-B/16 & 64.79 \\
        & CLIP ViT-B/32 & 65.10 \\
        \midrule
        \multirow{2}{*}{SSDL \cite{self_supervised}} & CLIP ViT-B/16 & 65.00 \\
        & CLIP RN-50 & 71.00 \\
        \midrule
        
        \multirow{3}{*}{\textbf{DPOD (Ours)}} & CLIP ViT-B/16 & \textbf{70.49} \\
        \textbf{DPOD (Ours)}& CLIP ViT-B/32 & \textbf{70.81} \\
          & CLIP RN-50 & \textbf{74.43} \\
        \bottomrule
    \end{tabular}
    }
    \end{adjustbox}
    \label{tab:acc_comp}
    \vspace{-10pt}
\end{table}

\section{Experimental Evaluation}
First, we evaluate the effectiveness of the proposed framework on the standard large-scale benchmark, namely NewsCLIPpings dataset~\cite{NewsCLIPpings}, which is created from VisualNews dataset~\cite{Visual_news}. 
It contains news articles  from sources like: The Guardian, BBC, USA Today, and The Washington Post, split into $71,072$ train, $7,024$ val and $7,264$ test examples. 
It contains news articles from 54 domains with varied amounts of training data across each of the domains, e.g., \textit{politics} (1396 samples), \textit{law-crime} (4348 samples), \textit{sports} (2786 samples) etc., reflecting the imbalance problem.



\noindent{{\bf Implementation Details.}}
We used CLIP ViT-B/32\cite{CLIP} model as the backbone for all our experiments. 
We use accuracy as the evaluation metric as in \cite{open_domain, self_supervised}, as the data has equal distribution of true and fake news.
Our codes are implemented in PyTorch and uses a single NVIDIA GeForce RTX 3080 Ti GPU.
ADAM optimizer with weight decay and stability constant is used for the training.
All the results are reported using the same set of parameters, e.g., $\beta=1.5$, $lr=1e-4$ and $batch size=64$ across all dataset splits. 
The values of hyperparameters are obtained during the validation process against the constant validation dataset containing all-domain data. 
These values were then kept fixed for the rest of the experiments.
Stage 1 is run for 40 epochs, which remains fixed even if the desired domain changes. We initialize our domain-agnostic prompts $\{V_{1}, V_{2}, V_{3}\}$ required in Stage 3, as 
``A photo of'' which is then learnt through training. After obtaining the output probability at the end of Stage 3, we use Sigmoid function to convert the output into a probability score $\hat{y}$ between 0 to 1. We train Stage 3 for 30 epochs. At the time of inference, the data having output greater than or equal to $0.5$ is assigned the label 1, i.e., Fake and the others are assigned label 0, i.e., True. 
Now, we describe the experiments conducted to evaluate the effectiveness and generalization of the proposed DPOD framework. 
We would specifically like to answer the following research questions: \\
i) {\em How does the model perform in the standard setting, where all the domain data is used for training and testing? \\
ii) How well does the proposed model perform for a desired target domain? \\
iii) Can our model generalize to completely unseen target domains and to other datasets? \\
iv) Can the model handle inconsistencies in domain labels and how well does it understand about the domains? \\
v) Are all the components in the model useful?} \\
\setlength{\tabcolsep}{4pt}
\begin{table}[t]
\caption{Comparison when the trained model is tested on target datasets having varying number of training samples (in parenthesis). Top (bottom) section contains results on domains having highest (lowest) number of samples.}
\centering
\scalebox{0.75}{
\begin{tabular}{llllll}
\toprule
\textbf{Model}        &  \multicolumn{5}{c}{\textbf{Maximum and Least Represented Domains}}                                                                                                                   \\ \midrule
        & \begin{tabular}[c]{@{}l@{}} Sport  \\ (3518) \end{tabular} & \begin{tabular}[c]{@{}l@{}}Law Crime \\ (4348)\end{tabular} & \begin{tabular}[c]{@{}l@{}}Internatn. \\ Relations  \\ (5832)\end{tabular} & \begin{tabular}[c]{@{}l@{}} World  \\ (5238) \end{tabular}  & \begin{tabular}[c]{@{}l@{}}Arts and\\ Culture  \\ (13736)\end{tabular} \\ \midrule
CLIP-FT\cite{open_domain} & 68.32 & 59.92                                                & 68.76                                                  & 62.11   & 68.52                                                      \\ 
CoOp\cite{coop}  & 69.68 & 62.87                                                & 69.13                                                  & 64.37   & 69.61                                                      \\

\textbf{DPOD (Ours)}    & \textbf{71.01} & \textbf{67.82}                                               & \textbf{69.47}                                                  & \textbf{65.24}   & \textbf{73.49}                                                      \\ \hline \hline
& \begin{tabular}[c]{@{}l@{}} Fashion  \\ (284) \end{tabular}& \begin{tabular}[c]{@{}l@{}}TV\\ Radio \\ (470)\end{tabular} & \begin{tabular}[c]{@{}l@{}}Education \\ (376) \end{tabular} & \begin{tabular}[c]{@{}l@{}} Money \\ (224) \end{tabular}  & \begin{tabular}[c]{@{}l@{}} Travel \\ (526) \end{tabular} \\ \midrule
CLIP-FT\cite{open_domain} & 56.68   & 59.90                                              & 63.22     & 61.19   & 59.48  \\
CoOp\cite{coop}  & 56.75   & 60.17                                              & 64.13     & 62.66   & 60.14  \\

\textbf{DPOD (Ours)}    & \textbf{58.92}   & \textbf{61.12}                                              & \textbf{65.07}     & \textbf{62.90}   & \textbf{62.70}  \\ 
\bottomrule
\end{tabular}
}
\label{tab:domains}
\vspace{-5pt}
\end{table}

\noindent
\textbf{Evaluation on standard setting for all domains:}
DPOD is a generalized framework, where the model is trained with data from all the available domains. 
First, we evaluate DPOD on the standard test protocol, where image-text pairs are available from all the domains, and report the results in Table \ref{tab:acc_comp}. 
Here, we assume that we know the domain label of the test data.
The experiment is run $5$ times with random sampling of training data, and the mean performance is reported. 
We compare our proposed framework with the previous SOTA approaches on NewsCLIPpings dataset, namely,
{\bf CLIP-FT}~\cite{open_domain},
{\bf SSDL~\cite{self_supervised}, DT-Transformer~\cite{papadopoulosacmmad}, MNSL\cite{zhangarxiv}, DCD\cite{huangicassp}, NewsCLIPpings\cite{NewsCLIPpings}}.
For CLIP-FT, the entire CLIP model is fine-tuned along with two additional layers with the available training data. 
This is the performance without using the open-domain data. 
For all the approaches, we directly report the accuracies from the respective papers.
When domain knowledge of the test sample is unknown, DPOD gives an accuracy of $69.13\%$, which is comparable to the performance achieved when the test domain information is known.
For fair comparison, we also perform our experiments with different variants of CLIP, namely, ViT-B/32, ViT-B/16 and RN-50. We observe that DPOD consistently outperforms the SOTA by a significant margin when similar backbones are used for comparison.

\begin{figure} 
    
        \centering
        \includegraphics[height=5cm]{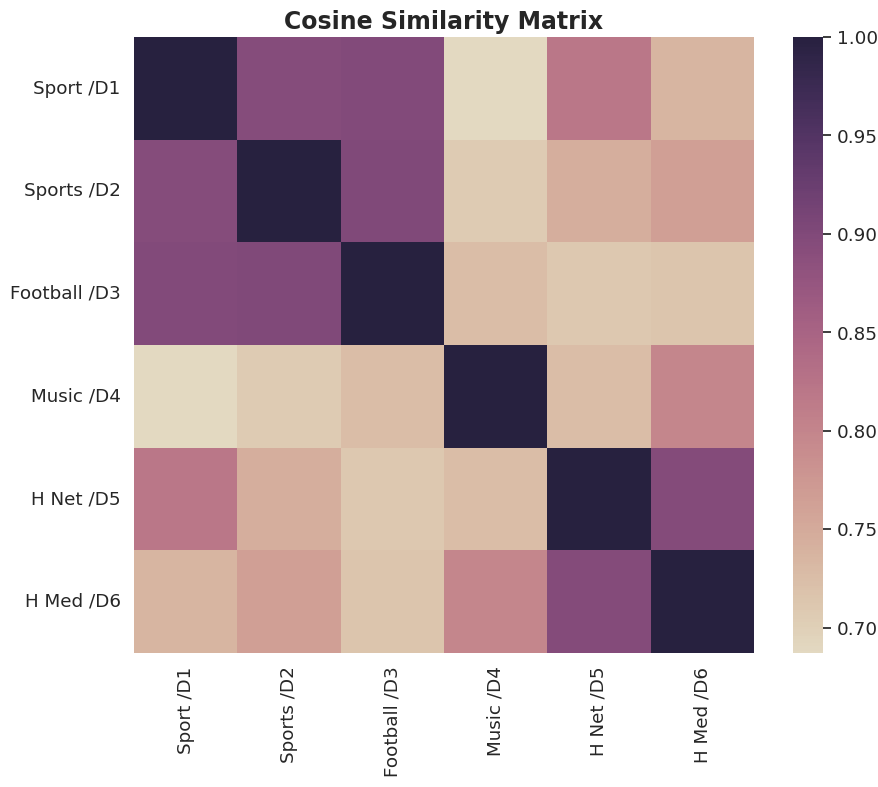}
        \caption{Similarity of the learnt domain-specific prompts of different domains. We observe that the sets \emph{Football, Sport, Sports} and \emph{Healthcare Network, Healthcare Medicine} have similar domain-specific prompts among themselves, but they have low similarity with each other and also with \emph{Music}.}
        \label{cos_sim}
    \vspace{-5pt}
\end{figure} 

\noindent
\textbf{Evaluation on a specific domain:}
Now, we consider the scenario where an individual is interested in news from a specific domain, eg a doctor might be interested in {\em healthcare} news.
Unlike existing methods, DPOD simultaneously specializes for each domain, by effectively utilizing out-of-domain samples.
To evaluate how the model handles data imbalance (from different domains) in the training data, we report results on $5$ domains with most datapoints and $5$ domains with the least data samples in Table \ref{tab:domains}.
In the absence of comparative approaches, we experiment with the following strong baselines:
i) {\bf Clip-FT} \cite{open_domain}: without the open-domain information for fair comparison with the other approaches. 
ii) {\bf CoOp}~\cite{coop}: First, the prompts $\{V_1, V_2, V_3\}$ are learnt using training data from all the domains. Then, these prompts are frozen for evaluation on the target domains. For fair comparison with our approach, CoOp is also used with A-CLIP.
All the models here are trained on the entire dataset, and then evaluated on the domain of interest.
We observe that DPOD consistently outperforms the other methods for all the domains, irrespective of the number of training examples available for that domain.

\begin{figure}[t!]
    \centering
    \includegraphics[height=4.3cm]{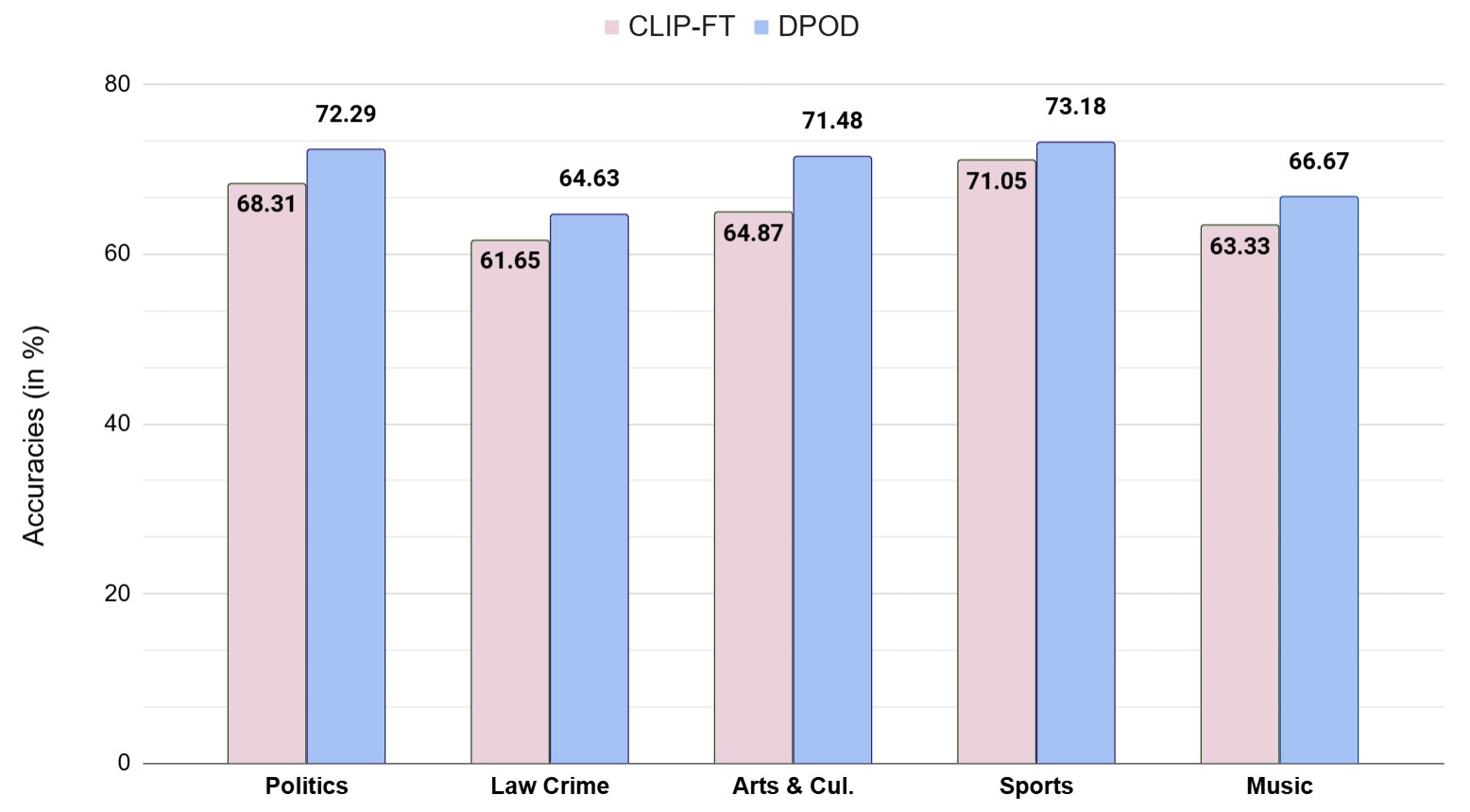}
    \caption{Comparison of accuracies (\%) where we train the model on 90\% domains and evaluate on the rest 10\% unseen domains.}
    \label{fig:unseen_domains}
\end{figure}

\vspace{0.1cm}
\noindent
\textbf{Generalization to unseen domains.}
In this dynamic world, the model may encounter test data, which it has not been trained on. 
Here, we evaluate the performance of our model when presented with completely new domains not included in the training set. For this, we train our model using samples from 90\% of the domains, and test it on the remaining 10\% unseen domains, using the most similar domain-specific prompt (based on cosine similarity). For a fair comparison, we applied a similar setting for CLIP-FT, and report the results in Fig. \ref{fig:unseen_domains}.
Here, the most similar domains corresponding to the 5 unseen domains obtained are as follows: \textit{politics}\textrightarrow\textit{intl relns}(sim=0.61), \textit{law crime}\textrightarrow\textit{conflict attack}(sim=0.78), \textit{arts cult.}\textrightarrow\textit{art \& design}(sim=0.59), \textit{music}\textrightarrow\textit{media}(sim=0.76), \textit{sports}\textrightarrow\textit{football}(sim=0.66).
Our proposed DPOD consistently outperforms the state-of-the-art CLIP-FT for all the unseen domain categories.

\begin{figure*}[t]
\centering
    {
    \includegraphics[width=\linewidth]{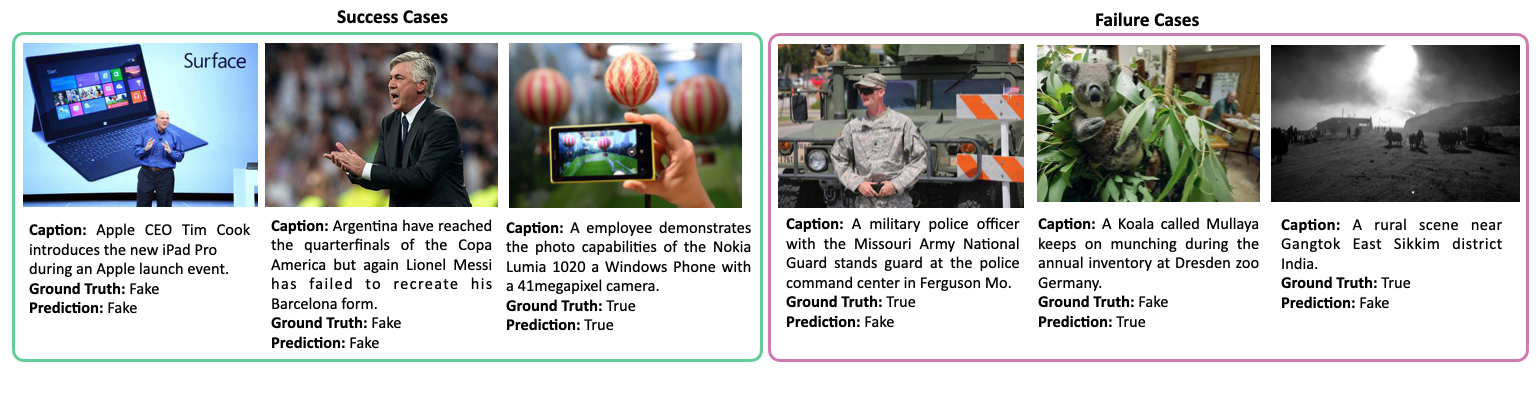}
    }
     \vspace{-18pt} 
    \caption{Qualitative results of the proposed DPOD Model. First three columns on the left with green border are examples of successful predictions, and the three on the right with pink border are examples of failure cases.}
    \label{fig:failsuccess}

\end{figure*}   

\begin{figure}[t]
\centering
    {
    \centering
    \includegraphics[width=\linewidth]{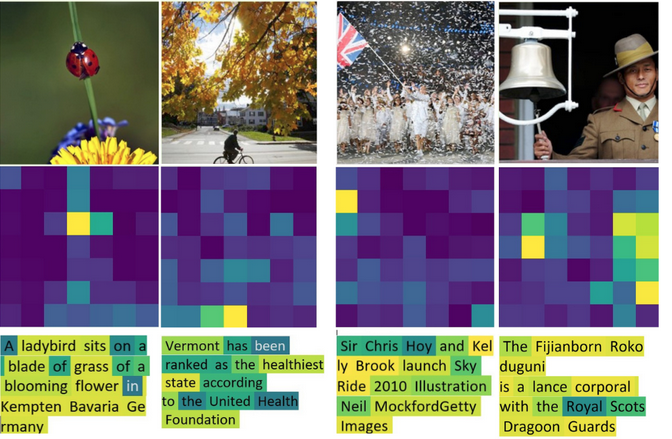}
    }
    \caption{The yellow regions show the portion of the image and text where A-CLIP attends to during classification.}
    \label{CLIPAttn}
    \vspace{-6pt} 
\end{figure}   


\begin{table}[t!]
\captionsetup{font=small}

  \centering
\footnotesize
 
  \begin{tabular}{ccc}
    \toprule
    \textbf{Stage 1} & \textbf{Stage 2} & \textbf{Accuracy(\%)} \\
    \midrule
    $\times$ & $\checkmark$ & $68.205$\\
    $\checkmark$ & $\times$ & $68.923$ \\
    \checkmark & \checkmark& $\textbf{70.811}$\\
    \bottomrule
  \end{tabular}
    \caption{Usefulness of the two stages of the proposed DPOD framework. Effectiveness of Stage 2 was evaluated by passing unique one-hot vectors corresponding to the unique domains.}
     \label{tab:ablation2}
    \vspace{-3pt} 
\end{table}



\begin{table}
\captionsetup{font=small}

\centering
\footnotesize
\begin{tabular}{lccc}
\toprule
Prompts &  \multicolumn{3}{c}{Test}\\
\midrule
         & Politics & Sport  &  All Domains  \\
\midrule
     \{${V}_{1}$, ${V}_{2}$, ${V}_{3}$\}           &   $77.027$      &       $69.680$ &  $69.349$     \\
     \{${V}_{1}$, ${V}_{2}$, ${V}_{3}$, ${V}_{4}$\}            &     $75.000$   &    $68.617$ & $70.351$ \\    
     {\bf DPOD}: \{${V}_{1}$, ${V}_{2}$, ${V}_{3}$, ${V}_{D}$\}        &    $\textbf{78.378}$    &     $\textbf{71.011}$ & $\textbf{70.811}$\\ 
\bottomrule
\end{tabular}
\caption{Usefulness of domain-specific prompt-tuning. Here the models have been trained on data from all the domains.}
\label{tab:ablation}
\end{table}

\vspace{0.1cm}
\noindent
\textbf{Handling Domain-Label Inconsistencies.}
Often, there are inconsistencies in the domain labels, which may be due to crowd-sourcing for obtaining labels.
For example, in the NewsCLIPpings dataset, there are domain labels like {\em football, sports, sport} and also {\em Healthcare Network, Healthcare Medicine}, etc. which should ideally have the same domain labels. 
The matrix in Fig.~\ref{cos_sim} shows the cosine similarity between the learnt domain-specific prompts of the corresponding two domains.
We observe that the learnt domain-specific prompts inherently captures the semantic information of the domains, thereby bringing the similar domains closer, thus helping in their training. 
The proposed DPOD does not require the exact domain names, and thus can work even if it is difficult to specify the domains, and dummy domain labels (like $D_1, D_2, etc.$) are used instead.

\vspace{0.1cm}
\noindent
\textbf{Cross Dataset Performance.} To further test the generalization capability of DPOD, we train our model and evaluate it on another real-world dataset, VERITE~\cite{verite}, which lacks domain information. For this, we create 10 clusters from the joint image-text embeddings using k-means, and create domain similarity vectors corresponding to these 10 cluster means. We then pass this through our trained model to create domain-specific vectors. We report the performances in Table \ref{tab:veritetab}. We observe that DPOD performs significantly better than the other methods in a similar setting.

\begin{table} [t]
\captionsetup{font=small}
\centering
\small
\begin{tabular}{lcc}

\toprule
\textbf{Method}        & \textbf{Backbone}      & \textbf{Accuracy (\%)} \\
\midrule
DT-Transformer~\cite{papadopoulosacmmad} & CLIP ViT-L/14 & $57.50$        \\
CLIP-FT~\cite{open_domain}        & CLIP ViT-B/32 & $56.78$        \\
\textbf{DPOD}  & CLIP ViT-B/32 & $\textbf{62.15}$ \\ 
\bottomrule
\end{tabular}
\caption{Cross dataset performance. All methods are trained on NewsCLIPpings and evaluated on VERITE. DPOD outperforms other methods including DT-Transformer of a superior backbone.}
\label{tab:veritetab}
\vspace{-5pt}
\end{table}
\vspace{0.1 cm}
\noindent
\textbf{Ablation Studies.}
Here, we analyze the importance of Stage 1 (label-aware alignment) and Stage 2 (semantic domain vectors) for the all domains scenario in Table 
 \ref{tab:ablation2}.
We observe that removing either of the stages deteriorates the overall performance. 
The second row in this table is generated by initializing $\{{V}_{1}, {V}_{2}, {V}_{3}\}$ with {\em ``A photo of''} and appending it with a fourth vector ${V}_{4}$, which is obtained by passing a unique one-hot vector corresponding to each training domains.
This implies that if the domains are treated independently, they cannot fully benefit from the training examples of closely related domains as in DPOD.
The importance of domain-specific prompts can also be observed from Table \ref{tab:ablation}, which shows the performance of
prompt  $\{{V}_{1}, {V}_{2}, {V}_{3}\}$  and $\{{V}_{1}, {V}_{2}, {V}_{3}, {V}_{4}\}$, where ${V}_{4}$ is another learnable prompt, but not a domain specific one as in DPOD (third row). Thus, all the proposed modules contribute to the improved performance of DPOD framework. 

\vspace{0.1cm}
\noindent
\textbf{Qualitative Results and Limitations.}
Here, we provide some qualitative results. 
Fig.~\ref{fig:failsuccess} (left with green border)  shows few examples of correct predictions.
We observe that the model shows the ability to accurately assess the veracity of a news item even in complex situations, as in the first example of a promotional event for a different brand.
Fig.~\ref{fig:failsuccess} (right with pink border) shows few examples of incorrect predictions by our model.
We observe that in some cases, the model struggles when multiple elements mentioned in the caption are present in the image, as {\em Koala} in the second failure case.
In Fig.~\ref{CLIPAttn}, we show some qualitative results of the A-CLIP attention maps on both the image and the text. We observe that, generally A-CLIP attends to regions (in yellow) that play a crucial role in decision making. 

\section{Conclusion}
Here, we proposed a novel framework DPOD for the challenging MFND task.
First, label-aware alignment of the data is achieved using the CLIP model to obtain generalized features. Further, we propose to learn generic as well as domain-specific prompts to classify the input image-text pairs. 
Experiments show that DPOD achieves the new state-of-the-art for this challenging, socially relevant MFND task.
\textbf{Acknowledgement.} This work was supported by DRDO and scientists from CAIR (DRDO) in particular.





%
%

{\small
\bibliographystyle{ieee_fullname}
\bibliography{egbib}

\begin{thebibliography}{10}\itemsep=-1pt

\bibitem{open_domain}
Sahar Abdelnabi, Rakibul Hasan, and Mario Fritz.
\newblock Open-domain, content-based, multi-modal fact-checking of out-of-context images via online resources.
\newblock In {\em {IEEE/CVF} Conference on Computer Vision and Pattern Recognition, {CVPR} 2022}, pages 14920--14929, 2022.

\bibitem{cosmos}
Shivangi Aneja, Chris Bregler, and Matthias Niessner.
\newblock Cosmos: Catching out-of-context image misuse using self-supervised learning.
\newblock In {\em Proceedings of the AAAI Conference on Artificial Intelligence}, volume~37, pages 14084--14092, 2023.

\bibitem{carte}
Benjamin Bowman, Alessandro Achille, Luca Zancato, Matthew Trager, Pramuditha Perera, Giovanni Paolini, and Stefano Soatto.
\newblock a-la-carte prompt tuning (apt): Combining distinct data via composable prompting.
\newblock In {\em Proceedings of the IEEE/CVF Conference on Computer Vision and Pattern Recognition}, pages 14984--14993, 2023.

\bibitem{simCLR_paper}
Ting Chen, Simon Kornblith, Mohammad Norouzi, and Geoffrey~E. Hinton.
\newblock A simple framework for contrastive learning of visual representations.
\newblock In {\em Proceedings of the 37th International Conference on Machine Learning, {ICML} 2020}, volume 119, pages 1597--1607, 2020.

\bibitem{bert}
Jacob Devlin, Ming-Wei Chang, Kenton Lee, and Kristina Toutanova.
\newblock Bert: Pre-training of deep bidirectional transformers for language understanding.
\newblock {\em arXiv preprint arXiv:1810.04805}, 2018.

\bibitem{poda}
Mohammad Fahes, Tuan-Hung Vu, Andrei Bursuc, Patrick P{\'e}rez, and Raoul de Charette.
\newblock Poda: Prompt-driven zero-shot domain adaptation.
\newblock In {\em Proceedings of the IEEE/CVF International Conference on Computer Vision}, pages 18623--18633, 2023.

\bibitem{CLIP-adapter}
Peng Gao, Shijie Geng, Renrui Zhang, Teli Ma, Rongyao Fang, Yongfeng Zhang, Hongsheng Li, and Yu Qiao.
\newblock Clip-adapter: Better vision-language models with feature adapters.
\newblock {\em CoRR}, abs/2110.04544, 2021.

\bibitem{dapl}
Chunjiang Ge, Rui Huang, Mixue Xie, Zihang Lai, Shiji Song, Shuang Li, and Gao Huang.
\newblock Domain adaptation via prompt learning.
\newblock {\em arXiv preprint arXiv:2202.06687}, 2022.

\bibitem{switchprompt}
Koustava Goswami, Lukas Lange, Jun Araki, and Heike Adel.
\newblock Switchprompt: Learning domain-specific gated soft prompts for classification in low-resource domains.
\newblock {\em arXiv preprint arXiv:2302.06868}, 2023.

\bibitem{huangicassp}
Mingzhen Huang, Shan Jia, Ming-Ching Chang, and Siwei Lyu.
\newblock Text-image de-contextualization detection using vision-language models.
\newblock In {\em ICASSP 2022-2022 IEEE International Conference on Acoustics, Speech and Signal Processing (ICASSP)}, pages 8967--8971. IEEE, 2022.

\bibitem{Allign}
Chao Jia, Yinfei Yang, Ye Xia, Yi-Ting Chen, Zarana Parekh, Hieu Pham, Quoc Le, Yun-Hsuan Sung, Zhen Li, and Tom Duerig.
\newblock Scaling up visual and vision-language representation learning with noisy text supervision.
\newblock In {\em International conference on machine learning}, pages 4904--4916. PMLR, 2021.

\bibitem{maple}
Muhammad~Uzair Khattak, Hanoona Rasheed, Muhammad Maaz, Salman Khan, and Fahad~Shahbaz Khan.
\newblock Maple: Multi-modal prompt learning.
\newblock In {\em Proceedings of the IEEE/CVF Conference on Computer Vision and Pattern Recognition}, pages 19113--19122, 2023.

\bibitem{supcon}
Prannay Khosla, Piotr Teterwak, Chen Wang, Aaron Sarna, Yonglong Tian, Phillip Isola, Aaron Maschinot, Ce Liu, and Dilip Krishnan.
\newblock Supervised contrastive learning.
\newblock {\em Advances in neural information processing systems}, 33:18661--18673, 2020.

\bibitem{Visual_news}
Fuxiao Liu, Yinghan Wang, Tianlu Wang, and Vicente Ordonez.
\newblock Visual news: Benchmark and challenges in news image captioning.
\newblock In Marie{-}Francine Moens, Xuanjing Huang, Lucia Specia, and Scott~Wen{-}tau Yih, editors, {\em Proceedings of the 2021 Conference on Empirical Methods in Natural Language Processing, {EMNLP} 2021}, pages 6761--6771. ACL, 2021.

\bibitem{NewsCLIPpings}
Grace Luo, Trevor Darrell, and Anna Rohrbach.
\newblock Newsclippings: Automatic generation of out-of-context multimodal media.
\newblock {\em CoRR}, abs/2104.05893, 2021.

\bibitem{self_supervised}
Michael Mu, Sreyasee~Das Bhattacharjee, and Junsong Yuan.
\newblock Self-supervised distilled learning for multi-modal misinformation identification.
\newblock In {\em {IEEE/CVF} Winter Conference on Applications of Computer Vision, {WACV} 2023}, pages 2818--2827, 2023.

\bibitem{papadopoulosacmmad}
Stefanos-Iordanis Papadopoulos, Christos Koutlis, Symeon Papadopoulos, and Panagiotis Petrantonakis.
\newblock Synthetic misinformers: Generating and combating multimodal misinformation.
\newblock In {\em Proceedings of the 2nd ACM International Workshop on Multimedia AI against Disinformation}, pages 36--44, 2023.

\bibitem{verite}
Stefanos-Iordanis Papadopoulos, Christos Koutlis, Symeon Papadopoulos, and Panagiotis~C Petrantonakis.
\newblock Verite: a robust benchmark for multimodal misinformation detection accounting for unimodal bias.
\newblock {\em International Journal of Multimedia Information Retrieval}, 13(1):4, 2024.

\bibitem{sniffer}
Peng Qi, Zehong Yan, Wynne Hsu, and Mong~Li Lee.
\newblock Sniffer: Multimodal large language model for explainable out-of-context misinformation detection.
\newblock In {\em Proceedings of the IEEE/CVF Conference on Computer Vision and Pattern Recognition}, pages 13052--13062, 2024.

\bibitem{CLIP}
Alec Radford, Jong~Wook Kim, Chris Hallacy, Aditya Ramesh, Gabriel Goh, Sandhini Agarwal, Girish Sastry, Amanda Askell, Pamela Mishkin, Jack Clark, Gretchen Krueger, and Ilya Sutskever.
\newblock Learning transferable visual models from natural language supervision.
\newblock In Marina Meila and Tong Zhang, editors, {\em Proceedings of the 38th International Conference on Machine Learning, {ICML} 2021}. {PMLR}, 2021.

\bibitem{cross_domain_FND}
Amila Silva, Ling Luo, Shanika Karunasekera, and Christopher Leckie.
\newblock Embracing domain differences in fake news: Cross-domain fake news detection using multi-modal data.
\newblock In {\em Thirty-Fifth {AAAI} Conference on Artificial Intelligence, {AAAI} 2021}, pages 557--565. {AAAI} Press, 2021.

\bibitem{vgg}
Karen Simonyan and Andrew Zisserman.
\newblock Very deep convolutional networks for large-scale image recognition.
\newblock {\em arXiv preprint arXiv:1409.1556}, 2014.

\bibitem{adclip}
Mainak Singha, Harsh Pal, Ankit Jha, and Biplab Banerjee.
\newblock Ad-clip: Adapting domains in prompt space using clip.
\newblock In {\em Proceedings of the IEEE/CVF International Conference on Computer Vision}, pages 4355--4364, 2023.

\bibitem{spotfake}
Shivangi Singhal, Rajiv~Ratn Shah, Tanmoy Chakraborty, Ponnurangam Kumaraguru, and Shin'ichi Satoh.
\newblock Spotfake: {A} multi-modal framework for fake news detection.
\newblock In {\em Fifth {IEEE} International Conference on Multimedia Big Data, BigMM 2019}, pages 39--47. {IEEE}, 2019.

\bibitem{EANN}
Yaqing Wang, Fenglong Ma, Zhiwei Jin, Ye Yuan, Guangxu Xun, Kishlay Jha, Lu Su, and Jing Gao.
\newblock {EANN:} event adversarial neural networks for multi-modal fake news detection.
\newblock In Yike Guo and Faisal Farooq, editors, {\em Proceedings of the 24th {ACM} {SIGKDD} International Conference on Knowledge Discovery {\&} Data Mining, {KDD} 2018}, pages 849--857. {ACM}, 2018.

\bibitem{Meta_neural_process}
Yaqing Wang, Fenglong Ma, Haoyu Wang, Kishlay Jha, and Jing Gao.
\newblock Multimodal emergent fake news detection via meta neural process networks.
\newblock In Feida Zhu, Beng~Chin Ooi, and Chunyan Miao, editors, {\em {KDD} '21: The 27th {ACM} {SIGKDD} Conference on Knowledge Discovery and Data Mining}, pages 3708--3716. {ACM}, 2021.

\bibitem{ecenet}
Fanrui Zhang, Jiawei Liu, Qiang Zhang, Esther Sun, Jingyi Xie, and Zheng-Jun Zha.
\newblock Ecenet: Explainable and context-enhanced network for muti-modal fact verification.
\newblock In {\em Proceedings of the 31st ACM International Conference on Multimedia}, pages 1231--1240, 2023.

\bibitem{zhangarxiv}
Yizhou Zhang, Loc Trinh, Defu Cao, Zijun Cui, and Yan Liu.
\newblock Detecting out-of-context multimodal misinformation with interpretable neural-symbolic model.
\newblock {\em arXiv preprint arXiv:2304.07633}, 2023.

\bibitem{dom_inv_prompt}
Cairong Zhao, Yubin Wang, Xinyang Jiang, Yifei Shen, Kaitao Song, Dongsheng Li, and Duoqian Miao.
\newblock Learning domain invariant prompt for vision-language models.
\newblock {\em arXiv preprint arXiv:2212.04196}, 2022.

\bibitem{cocoop}
Kaiyang Zhou, Jingkang Yang, Chen~Change Loy, and Ziwei Liu.
\newblock Conditional prompt learning for vision-language models.
\newblock In {\em Proceedings of the IEEE/CVF Conference on Computer Vision and Pattern Recognition}, pages 16816--16825, 2022.

\bibitem{coop}
Kaiyang Zhou, Jingkang Yang, Chen~Change Loy, and Ziwei Liu.
\newblock Learning to prompt for vision-language models.
\newblock {\em International Journal of Computer Vision}, 130(9):2337--2348, 2022.

\bibitem{SAFE}
Xinyi Zhou, Jindi Wu, and Reza Zafarani.
\newblock {SAFE:} similarity-aware multi-modal fake news detection.
\newblock In {\em Pacific-Asia Conference on knowledge discovery and data mining}, pages 354--367. Springer, 2020.

\bibitem{CLIP_guided}
Yangming Zhou, Qichao Ying, Zhenxing Qian, Sheng Li, and Xinpeng Zhang.
\newblock Multimodal fake news detection via clip-guided learning.
\newblock {\em CoRR}, abs/2205.14304, 2022.

\end{thebibliography}
}

\end{document}